\documentclass[letterpaper, 10 pt, conference]{ieeeconf}  % Comment this line out if you need a4paper

\IEEEoverridecommandlockouts                              % This command is only needed if 
\usepackage{amssymb}
\usepackage{graphicx}
\usepackage{wrapfig}
\usepackage{booktabs}
\usepackage{multirow}
\usepackage{caption}
\usepackage{array}  % for better column formatting
\usepackage{arydshln}
\usepackage{siunitx}
\usepackage{algorithm}
\usepackage{algpseudocode}
\usepackage{wrapfig}
\usepackage{xcolor}
\usepackage{amsmath}
\usepackage{makecell}
\usepackage{cite}
\usepackage[normalem]{ulem}
\usepackage{url}
\usepackage{titlesec}
\usepackage[font=footnotesize]{caption}

\titlespacing{\section}{0pt}{*1.0}{*0.5}
\titlespacing{\subsection}{0pt}{*0.8}{*0.4}

\newcommand{\pose}{\ensuremath{x}}
\newcommand{\poses}{\ensuremath{\mathcal{X}}}
\newcommand{\terrainFunc}{\ensuremath{T}}

\newcommand{\terrains}{\ensuremath{\mathcal{T}}}

\newcommand{\strength}{\ensuremath{\alpha}}

\newcommand{\lm}[1]{\textcolor{black}{#1}}
\newcommand{\ourapproach}{\textsc{spacer}}
\newcommand{\regret}{\ensuremath{\rho}}

\title{Terrain Costmap Generation via Scaled Preference Conditioning}

\author{Luisa Mao$^{1}$, Garrett Warnell$^{1,2}$, Peter Stone$^{1,3}$, Joydeep Biswas$^{1,4}$% <-this % stops a space
\vspace{-5mm}
\thanks{\scriptsize $^{1}$ The University of Texas at Austin, Austin, TX, USA.
        }%
\thanks{\scriptsize $^{2}$ DEVCOM Army Research Laboratory, Austin, TX, USA.}%
\thanks{\scriptsize $^{3}$ Sony AI,  Boston, MA 02129, USA.
       }%
\thanks{\scriptsize $^{4}$ NVIDIA, Santa Clara, CA 95051, USA.
        }%
}

\begin{document}
\maketitle

%===============================================================================

\begin{abstract}
    Successful autonomous robot navigation in off-road domains requires the ability to generate high-quality terrain costmaps that are able to both generalize well over a wide variety of terrains and rapidly adapt relative costs at test time to meet mission-specific needs. Existing approaches for costmap generation allow for either rapid test-time adaptation of relative costs (e.g., semantic segmentation methods) or generalization to new terrain types (e.g., representation learning methods), but not both. In this work, we present \textit{scaled preference conditioned all-terrain costmap generation} (\ourapproach{}), a novel approach for generating terrain costmaps that leverages synthetic data during training in order to generalize well to new terrains, and allows for rapid test-time adaptation of relative costs by conditioning on a user-specified \textit{scaled preference context}. Using large-scale aerial maps, we provide empirical evidence that \ourapproach{} outperforms other approaches at generating costmaps for terrain navigation, with the lowest measured \textit{regret} across varied preferences in five of seven environments for global path planning.

    % \lm{tldr: We introduce a new method for terrain costmap generation that combines strong generalization to novel terrains with rapid test-time adaptability to scaled user preferences, using a unique preference formulation and training entirely on synthetic data.}

\end{abstract}

%===============================================================================

% \lm{story}
% \begin{itemize}
%     \item \textbf{Problem}. Want a method of terrain costing where humans have control over the magnitude of the costs that can work in zero-shot settings \lm{clarify what zero-shot setting is}.
%     \item \textbf{Classic semantic segmentation.} Locked to a specific ontology and therefore cannot generalize to zero-shot settings.
%     \item \textbf{Open-set semantic segmentation.} Have encoded priors over terrain appearance and therefore cannot generalize OOD (eg, blue football field is still grass, bad performance of Clip Seg in hard environments, NIR imagery).
%     \item \textbf{representation learning.} todo
%     \item \textbf{PACER.} Humans do not have control over magnitude of the resulting costs.
%     \item \textbf{PACER2.} Generalizes to zero-shot terrains and provides method of controlling magnitude of resulting costs.
% \end{itemize}

% Pacer 2 features
% more expressive with respect to terrains
%     \begin{itemize}
%         \item vs pacer, longer contexts. pacer 1 was limited by the length of the context and so needed a realistic prior/data to be able to interpolate preferences. Pacer 2 no longer needs this since we have a variable context length (show in figure in ablation study). Also has strengths.

%         \item vs semantics, provide visuals and not limited by words
%     \end{itemize}

% \lm{we find that ClipSeg is the best foundation model for segmenting aerial imagery, better than grounded SAM, representative of that family of models}

\section{Introduction}

\lm{Off-road navigation requires terrain understanding that is both flexible and robust. Robots must (1) recognize diverse, even novel or mixed terrains, and (2) adapt traversal costs at test time for mission goals. Existing costmap methods struggle to achieve both generalization and adaptability simultaneously.}

Approaches based on semantic segmentation support test-time variation of relative costs by assigning labels to terrain types and allowing users to directly adjust cost values whenever they wish. However, these methods are typically restricted to a fixed, predefined set of terrain classes, limiting their effectiveness in unfamiliar or ambiguous environments \cite{jiang2021rellis, guan2022ga, luddecke2022image}. Conversely, representation learning approaches operate over feature spaces that allow them to better generalize to out-of-distribution inputs, but they typically lack mechanisms for easily adjusting relative costs at test time, making it difficult for users to express preferences or adapt to shifting objectives mid-mission \cite{yao2022rca,karnan2023sterling, sikand2022visual}.

\begin{figure}[h]
    \centering
    \includegraphics[width=.4\textwidth]{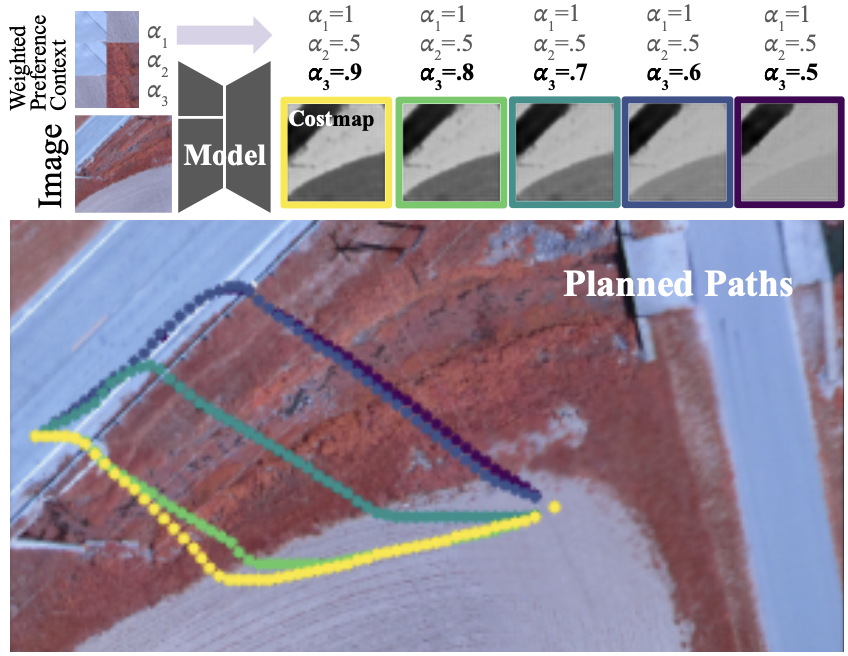}  
    \caption{Given an input image and preference over terrains, the model outputs a costmap. A scaled preference context is a set of pair comparisons of terrains (left patch preferred over right), scaled by a \textit{strength of preference} $\alpha$. Varying the strength of a preference leads to different gradation of costs and resulting planned paths.}
    \label{fig:overview_fig}
    \vspace{-3mm}
\end{figure}

In this work, we introduce \textit{scaled preference conditioned all-terrain costmap generation} (\ourapproach{}), a novel method for terrain costmap generation that uniquely supports user-specified preferences over terrain pairs (Fig. \ref{fig:overview_fig}). \ourapproach{} is conditioned on a novel, flexible, and expressive \textit{scaled preference context}—a variable-length list of terrain image pairs annotated with a numerical preference strength—enabling nuanced, test-time control over terrain costs. \lm{Unlike prevalent preference-based inverse reinforcement learning (PbIRL) methods, which learn a reward function offline, our approach learns \textit{how} to perform PbIRL at runtime. During inference, the model receives only a small number of preferences, using them as contextual cues to infer mission-specific reward structures. This capability emerges from training on a wide variety of preference types and their corresponding reward functions, enabling the model to relate new preferences to appropriate reward functions at test time.} Our method allows human operators to specify any number of scaled preferences over pairs of terrain patches by simply clicking on regions of an image and specifying the strength of preference for that pair. In addition to this input modality, \ourapproach{} generalizes well to novel and complex terrains by training on a large, diverse set of synthetic terrains and leveraging internet-pretrained visual representations. \lm{For example, if a last-mile delivery robot struggles with tall grass, this method enables rapid preference adjustments toward low grass or pavement, eliminating the need for retraining.}

We evaluate \ourapproach{} across a range of aerial environments and preferences, and demonstrate that it consistently outperforms existing methods at generating mission-aligned costmaps. To further validate generalization and real-world applicability, we conduct both quantitative and qualitative evaluations on the RELLIS-3D dataset  \cite{jiang2021rellis}—a high-resolution, multimodal benchmark for off-road, unstructured environments with diverse terrain types and fine-grained semantic labels. Additionally, we present ablation studies that isolate the impact of key architectural and training design choices.

\section{Previous Work}
Our work sits at the intersection of terrain-aware navigation and preference-based inverse reinforcement learning (PbIRL). Below, we review previous work in each area.

\subsection{Terrain-Aware Navigation via Costmaps} 

Our method transforms terrain images into costmaps usable by planners for terrain-aware navigation. Existing approaches generally fall into two categories: those using semantic representations and those using learned embeddings of terrain appearance.

Semantic approaches, such as classical or open-set semantic segmentation \cite{Meng-RSS-23, guan2022ga, luddecke2022image}, assign terrain labels and then map these labels to user-defined costs. While this enables rapid test-time cost adjustment, classical methods are limited to predefined terrain classes, and open-set variants remain biased toward visually canonical terrains (e.g., green grass, gray concrete), which may fail in unfamiliar environments.

Embedding-based methods instead represent terrain as points in a learned feature space, removing the need for fixed semantic labels. However, because this space is not directly interpretable, cost functions must be learned through demonstrations or preferences. Most such methods require retraining whenever user preferences change \cite{karnan2023sterling, yao2022rca, zurn2020self, sikand2022visual, zhang2025creste} or restrict how relative costs can be tuned \cite{pacer}.

\subsection{Preference-based Inverse Reinforcement Learning}

Our approach builds on the framework of preference-based inverse reinforcement learning (PbIRL) \cite{Wirth2017}, which learns reward (or cost) functions from user preferences. Traditional PbIRL methods learn these reward functions offline from large, fixed datasets and are applied in domains such as video games \cite{Christiano2017} and robotics \cite{Schoenauer2014}. In contrast, \ourapproach{} performs PbIRL at runtime: during inference, the model receives only a small number of scaled preferences and uses them as contextual cues to infer a task-specific reward function. This ability arises from training on a wide variety of preference types and their corresponding reward functions, enabling generalization from few preferences at test time.

Closest to our work, \cite{Zucker2010} and \cite{pacer} also infer terrain costs from user preferences over terrain pairs, but they do not model preference scale. While prior work such as \cite{Ghosal2023} encodes varying feedback noise through the rationality coefficient of a Boltzmann model, our scalar instead expresses preference certainty, producing stronger cost separation when the user expresses higher confidence.

\lm{\subsection{Summary and Motivation}
Existing approaches do not support both \textbf{deployment in unseen environments} and \textbf{test-time modulation of terrain costs according to user preferences}—particularly those expressed with scalable strength. \textsc{spacer} addresses both limitations.}

%===============================================================================

\section{\lm{Problem:} Preference-Aligned Path Planning}
\lm{In this section, we formalize the preference-aligned path planning problem as introduced by prior work, which serves as the problem setting for our method described in Sec. \ref{sec:method}}. We focus on the SE(2) path planning problem, where the goal is to find a trajectory of poses $\Gamma = \{\pose_1, \ldots, \pose_n\}$, $\pose_i \in \poses{}$, that minimizes a cost function $C$:

\[
\Gamma^* = \arg\min_{\Gamma} \sum_i C(\Gamma_i)
\quad \text{s.t.} \quad 
\begin{aligned}
& \pose{}_1 = \pose{}_{\mathrm{start}}, \\
& \pose{}_n = \pose{}_{\mathrm{goal}}, \\
& \|\pose{}_{i+1} - \pose{}_i\| \leq \delta, \; \forall i.
\end{aligned}
\] where $\delta$ is a maximum step length determined by the discretization or dynamics.

In the terrain-aware preference-aligned setting, $C$ is shaped by \lm{a black-box} human preference function $H$ over terrains $\tau \in \mathcal{T}$, mapping each terrain to a non-negative scalar cost $H(\tau) \in \mathbb{R}^{0+}$. These preferences capture subjective factors—e.g., some terrains may be physically damaging to the robot or undesirable for social or aesthetic reasons. Additionally, the robot does not directly observe terrains but receives visual input via an observation function $O(\poses{}, \terrainFunc{}) = I$ that produces an image from a given pose and the global terrain map $\terrainFunc{}$. Prior work, \textsc{pacer} \cite{pacer}, introduced a costmap generator $R$ that implicitly models $H$ such that costmap $C = R(O(x,T)|H)$, and provided a method for approximating $R$ and $H$, where \textbf{$H$ and $\terrains{}$ could be unseen during training}. 
% \textsc{pacer} showed that a \textit{necessary} condition to model $H$ was to preserve the ordering of terrains according to preferences, but could not recover relative cost magnitudes, which are also important for planning.

\lm{\textsc{pacer} showed that a \textit{necessary} condition to model $H$ was to preserve the ordering of terrains according to preferences. Let $x_\tau$ denote a pose on terrain $\tau$ and $C|_x \in \mathbb{R}^{0+}$ the costmap evaluated at pose $x$. Given preference  $\tau_a \succ \tau_b$ from $H$ denoting terrain $\tau_a$ is preferred to $\tau_a$, the cost at $x_{\tau_b}$ is less than that at $x_{\tau_b}$:
\begin{align}
\text{if } \tau_a \succ \tau_b, \quad \tau_a, \tau_b \in \terrains{} \quad  \text{then } \quad 
C \big|_{x_{\tau_a}} 
< C \big|_{x_{\tau_b}}
\end{align}}

% \lm{move notation above the equation. alphas only denote pairwise strength. explain more--spell it out}

\lm{However, this condition doesn't capture the magnitude of preferences, which are also important for planning. In this work, we aim not only to recover preferences but also to preserve their strength in the resulting cost estimates. Let $\tau_a \succ^{\alpha_{ab}} \tau_b$ denote a preference weighted by $\alpha_{ab} \in [0, 1]$. A small $\alpha_{ab}$ (i.e. a weak preference) should result in a smaller cost difference in $R$ conditioned on $H$  and stronger preferences should yield larger differences:}

% That is, a pair of terrains with a stronger preference signal in
% $H$—i.e., a larger margin between their values—should result in a larger cost difference in $R$ conditioned on $H$ and weaker preferences should yield smaller differences:
\lm{
\vspace{-1mm}
\begin{align}
\text{if } & \tau_a \succ^{\alpha_{ab}} \tau_b,\;
  \tau_c \succ^{\alpha_{cd}} \tau_d,\;
  \alpha_{ab} < \alpha_{cd};\;
  \tau_{a,b,c,d} \in \mathcal{T},
\nonumber \\
\text{then } &
\| C|x_{\tau_a} - C|x_{\tau_b} \|
< \| C|x_{\tau_c} - C|x_{\tau_d} \|. \label{eq:condition}
\end{align}
 % where $\tau_a \succ^{\alpha_{ab}} \tau_b$ denotes preference weighted by a scalar $\alpha_{ab}$. 
 Thus, if the human shows a much stronger preference between the terrains at poses $\pose_{\tau_c}$ and $\pose_{\tau_d}$ than between those at $\pose_{\tau_a}$ and $\pose_{\tau_b}$, then the costmap $C = R(O(\cdot, T) \mid H)$ should reflect this difference by assigning a correspondingly larger cost difference between $\pose_{\tau_c}$and $\pose_{\tau_d}$ than $\pose_{\tau_a}$ and $\pose_{\tau_b}$.}

% if $\tau_a \succ \tau_b$ then $R(O(x, T) | H)|x_{\tau_a} < R(O(x, T) | H)|x_{\tau_b} $. where ... ... but this doesn't say anything abt distance between the costs... so better is (in addition to the condition above, we now have the one below)
% if $\tau_a \succ \tau_b$ with strength $\alpha_{ab}$ and $\tau_c \succ \tau_d$ with $\alpha_{cd}$ where $\alpha_{ab} < \alpha_{cd}$ then $||R(O(x, T) | H)|x_{\tau_a} - R(O(x, T) | H)|x_{\tau_b} || < ||R(O(x, T) | H)|x_{\tau_c} - R(O(x, T) | H)|x_{\tau_d} ||$

% In this work, we aim not only to recover preferences but also to preserve the relative strength of those preferences in the resulting cost estimates. That is, a pair of terrains with a stronger preference signal in
% $H$—i.e., a larger margin between their values—should result in a larger cost difference in $R$ conditioned on $H$.
% In contrast, weaker preferences should yield smaller differences. Thus, if the human shows a much stronger preference between the terrains at poses $\pose_1$ and $\pose_2$ than between those at $\pose_3$ and $\pose_4$, then the costmap $C = R(O(\cdot, T) \mid H)$ should reflect this difference by assigning a correspondingly larger cost difference between $\pose_1$ and $\pose_2$ than between $\pose_3$ and $\pose_4$, and vice versa.

\lm{This section reviewed the preference-aligned path planning framework that motivates our work. Section IV presents our proposed costmap generation method, followed by dataset curation and training details in Section V.}

\section{\lm{Method:} Preference-Aligned All-Terrain Costmap Generation}
\label{sec:method}

We now describe \textit{scaled preference conditioned all-terrain costmap generation} (\ourapproach{}), our approach to terrain costmap generation that both \textbf{generalizes well to new types of terrain} and also allows for \textbf{rapid test-time adaptation to new relative costs}\lm{, which builds upon the problem formulation introduced in Section III.}
First, we define and connect to the notion of terrain cost a \textit{scaled preference context}, a novel form of human input that allows users to specify not only their preferences between two options but also a scalar-valued weight indicating the \textit{strength} of that preference.
Next, we motivate and describe the neural network architecture that we adopted for \ourapproach{}.

    \subsection{Scaled Preference Context}

    We ask users to provide a \textit{scaled preference}, which we define to be both a preferred option between to alternatives (here, terrains) and also a scalar-valued \textit{strength} of that preference, where a larger strength communicates a stronger preference for the preferred option.
    More formally, we define a scaled preference as a tuple $\xi = (\tau^{(1)}, \tau^{(2)}, \alpha)$, which denotes that terrain $\tau^{(1)}$ is preferred to $\tau^{(2)}$ (lower costs as more preferred) with strength $\alpha \in [0, 1]$.

    To model the underlying terrain costs from a scaled preference $\xi$, we employ the Bradley–Terry model \cite{bradley_terry}, a well-established framework for reasoning over pairwise comparisons. The forward Bradley–Terry model relates pairwise match probabilities to costs, while the inverse model infers latent costs from given probabilities. It is suitable in our case because it allows us to infer the hidden human cost function $H$ consistent with given preference strengths \strength{}.

    Given the user-provided strength $\alpha$ of how much $\tau^{(1)}$ is preferred to $\tau^{(2)}$, the \textit{inverse} Bradley-Terry model can be used to infer the costs $H(\tau^{(2)})$ and $H(\tau^{(1)})$ assigned by the hidden human preference function. Since $\tau^{(1)}$ is preferred to $\tau^{(2)}$ as given in the ordered tuple $\xi$, the probability $P(H(\tau^{(2)}) > H(\tau^{(1)}))$ is in [0.5, 1] so we map the strength $\alpha \in [0,1]$ to $[0.5, 1]$ when converting it to a probability.
    \textbf{Succinctly, we interpret $\frac{\alpha+1}{2}$ to be the probability $P(H(\tau^{(2)}) > H(\tau^{(1)}))$}. The strength \strength{} and $H$ are related by the \textit{forward} Bradley Terry model, as shown below:

    \begin{align}
        \frac{\alpha + 1}{2}
        & = \frac{\exp\{H(\tau^{(2)})\}}{\exp\{H(\tau^{(1)})\} + \exp\{H(\tau^{(2)})\}} \\
        &= \sigma(H(\tau^{(2)}) - H(\tau^{(1)}))
    \end{align}

    Thus, given a sample $\xi = (\tau^{(1)}, \tau^{(2)}, \alpha)$, the \textit{inverse} Bradley–Terry model can be applied to recover an estimate hidden preference function $H$. \lm{\textbf{Relating to condition \ref{eq:condition}, larger user-supplied $\alpha$s lead to larger differences in costs.}}
   
    In this framework, the strength parameter \strength{} acts as a control variable, allowing the user to modulate the relative cost difference between terrains in the inferred model. This formulation ensures that stronger preferences correspond to larger cost differences in the recovered latent function \( H \), and provides a method for recovering terrain costs from human preference data.  We refer to an unordered collection of $K$ scaled preferences as a \textit{scaled preference context} $\Xi = (\xi_1, \ldots, \xi_K)$.

    Therefore, our approach to predicting costmap $\hat{C}$ is a conditional inference problem given by

    \begin{equation}
    {R}( I | H) \approx \hat{R}_\phi( I | \Xi) = \hat{C}
    \end{equation} 
    where $\hat{R}_\phi$ is a neural network parameterized by $\phi$, which implicitly learns the inverse Bradley Terry model given $\Xi$.

\subsection{Model Architecture}
\begin{figure}[h]  % 'h' places the figure approximately here
    \centering
    \includegraphics[width=.5\textwidth]{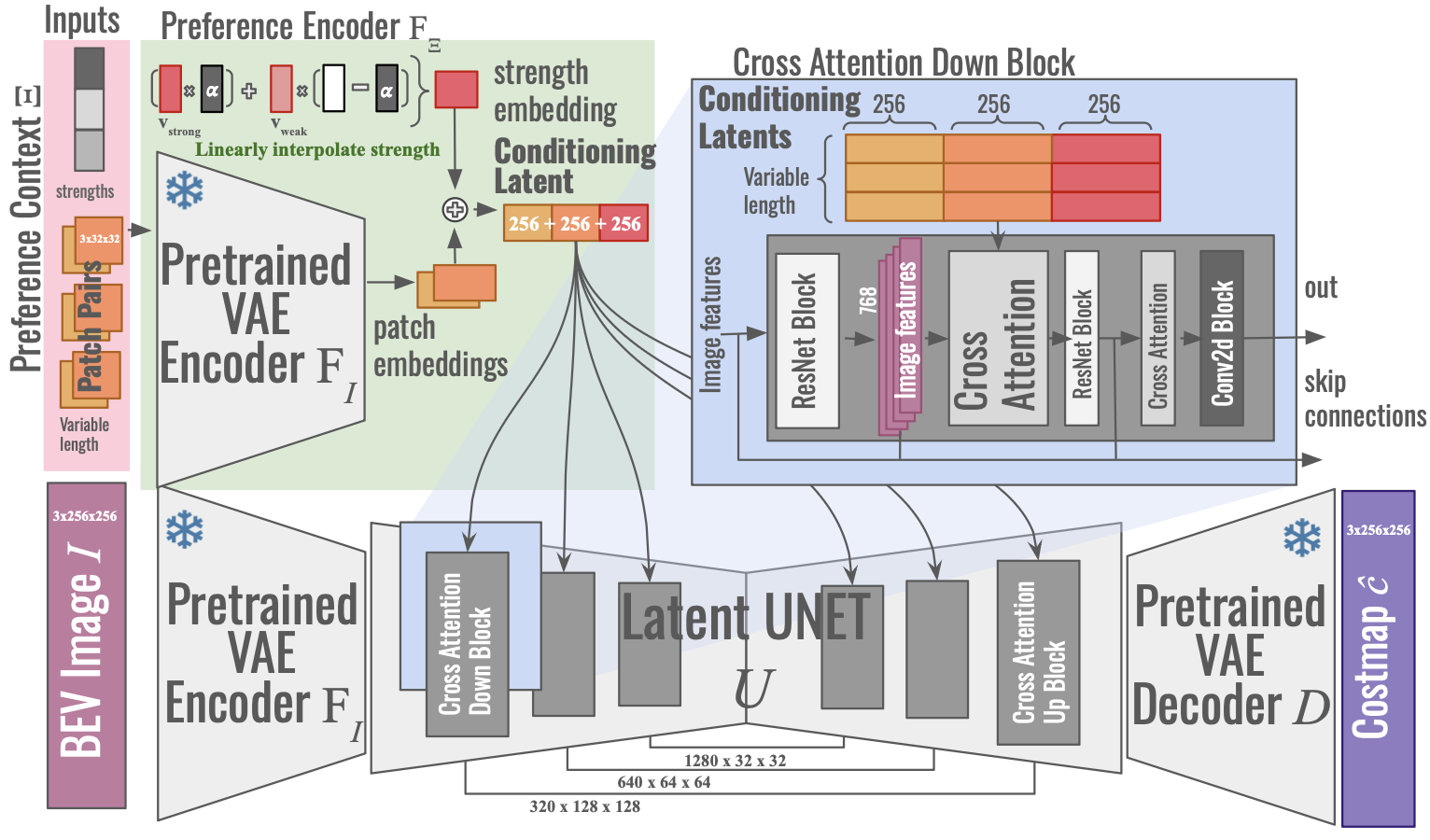}  % Replace with your filename
    \caption{Model architecture. Inputs are BEV image and variable-length preference context consisting of patch pairs and strengths. The conditioning latents are the concatentation of patch embeddings and strength embedding (calculated as a linear interpolation of \textit{strong} and \textit{weak} embeddings). The output is a costmap. \lm{updated}}
    \label{fig:model_arch}
    \vspace{-3mm}
\end{figure}

    We model the terrain cost function as a neural network that maps a terrain image to a costmap conditioned on the scaled preference context (Fig. \ref{fig:model_arch}).
    Our model consists of a \textit{scaled preference context encoder} $F_{\Xi}$, image encoder $F_{I}$, latent UNet $U$, and decoder $D$. We give a novel interpretation of preference strength as an interpolation in a latent space, with $\alpha$ controlling the tradeoff between $v_{strong}$ and $v_{weak}$ embeddings. $v_{strong}$ indicates the preference in the ordered pair is strong, whereas $v_{weak}$ indicates no preference. The model formulation is presented in equation \ref{eq:model_formulation}. Equation \ref{eq:pref_context_encoder} shows the construction of the \textit{scaled preference context} embeddings.
    
    \textbf{Preference Context Encoder}: The preference context encoder $F_\Xi$ maps each terrain image patch in $\Xi$ into the latent space of the image encoder $F_I$.  
    For each tuple $(\tau_i^{(1)}, \tau_i^{(2)}, \alpha_i)$, both terrain patches are encoded with $F_I$, and the preference strength $\alpha_i$ is embedded as a weighted combination of $v_{\text{strong}}$ and $v_{\text{weak}}$:
    \begin{equation}
    \label{eq:pref_context_encoder}
    \begin{aligned}
    F_\Xi(\Xi) &= \big[ F_I(\tau_i^{(1)}),\; F_I(\tau_i^{(2)}),\; \alpha_i v_{\text{strong}} + (1 - \alpha_i)v_{\text{weak}} \big]_{i=1}^K
    \end{aligned}
    \end{equation}

    The injection of the preference strength 
    \strength{} into the network is a core innovation of our architecture. Instead of passing \strength{} directly as a scalar, we map it into a latent embedding space via a convex combination of two learned embeddings, $v_{\mathrm{strong}}$ and  $v_{\mathrm{weak}}$, denoting strong and neutral preferences: $v_{\mathrm{strong}} \cdot (\strength{}) + v_{\mathrm{weak}} \cdot (1-\strength{})$.  Our design is inspired by the time embeddings used in diffusion models \cite{stable_diffusion_Rombach_2022_CVPR}, which have proven effective for conditioning on continuous variables. The patch latents are concatenated channel-wise with this \textit{strength embedding}, providing the model with a rich conditioning signal.

      \textbf{Pretrained Encoder and Decoder}: The costmap $\hat{C}$ is predicted from the input image $I$ and the preference context $\Xi$. We use $F_I$ and $D$ as the pretrained VAE encoder and decoder from Stable Diffusion \cite{stable_diffusion_Rombach_2022_CVPR}, with weights frozen during training, and $U$ as a latent conditional UNet:
    \begin{equation}
    \label{eq:model_formulation}
    \begin{aligned}
    \hat{C} &= \hat{R}_\phi(I \mid \Xi) \\
           &= D\Big( U \big[
                F_I(I),\;
                F_\Xi(\Xi) \big]
              \Big)
    \end{aligned}
    \end{equation}

\textbf{Latent UNet:} The latent UNet incorporates the conditioning embedding with the image embedding through cross attention layers, so the preference context may have a variable length. The outputs of the UNet are latent vectors which are decoded by the frozen VAE decoder $D$ into a local BEV costmap. The mean of the three output channels is interpreted as the costmap $\hat{C}$.

\section{Dataset Curation and Training}
\lm{Since collecting large-scale human preference data is challenging, we procedurally synthesize a dataset consistent with our preference model to enable scalable training (further described in Sec. \ref{sec:synth_dataset}).}
Our dataset $\mathcal{D} = \{(I_i, \Xi_i, S_i, C_{T_i})\}_{i = 1}^n$ consists of an image $I$, scaled preference context $\Xi$, list of segmentation masks $S$, and target costmap $C_T$. Terrain classes are only used during training. We do not require a global set of terrain classes during deployment.
\vspace{0mm}

\subsection{Loss Function}

Our objective is to learn a costmap prediction $\hat{C}$ that (1) is consistent with human preferences over terrain and (2) well-regularized. 

First, by $\mathcal{L}_1$, the model outputs are encouraged to match the given preference context $\Xi$. To extract the costs of $k$ terrain classes, we use segmentation masks $S = \{s_1, \dots, s_k\}$, where $\mu_s(C)$ denotes the mean over mask $s$ of costmap $C$. By the forward Bradley-Terry model, we match the log-odds calculated from the predicted costmap $\mu_{s_i^{(1)}}(\hat{C}) - \mu_{s_i^{(2)}}(\hat{C})$ to the target log-odds given in the preference context 
$\sigma^{-1}(\frac{\strength{}_i+1}{2})$ to ensure predicted costs are consistent with the user-provided preference.

Second, by $\mathcal{L}_2$, we introduce an additional term that anchors the predicted costs to a target distribution, ensuring they remain within the range $(0, 1)$ (see Section~\ref{sec:ablations} for ablation results). $\mathcal{L}_2$ prevents drift, since relative preferences are invariant under a uniform scaling of costs, optimizing only for preference consistency can cause the learned costmap to drift in magnitude. 

For a given scaled preference context $\Xi$, predicted costmap $\hat{C}$, and target costmap $C_T$, we define the loss as: 
\vspace{-1mm}
\begin{align}
    \mathcal{L} &= \mathcal{L}_1 + \lambda \mathcal{L}_2, \\
    \mathcal{L}_1 &= \sum_{i \leq |\Xi|} \mathcal{L}_\delta\!\left( \mu_{{s_i^{(2)}}}(\hat{C}) - \mu_{s_i^{(1)}}(\hat{C}), \; \sigma^{-1}\!\left(\frac{\alpha_i + 1}{2}\right) \right), \label{eq:loss1} \\
    \mathcal{L}_2 &= \mathcal{L}_\delta(\hat{C}, C_T), \label{eq:loss2}
\end{align}
where $\lambda$ tunes the relative strength between $\mathcal{L}_1$ and $ \mathcal{L}_2$, $\mathcal{L}_\delta$ is the Huber loss, and $\sigma^{-1}$ is the inverse sigmoid. We additionally train with a low learning rate of 1e-5 and gradient accumulation. 
Fig. \ref{fig:training_example_loss_function} shows the loss function.

    \subsection{Data Curation}
    \label{sec:synth_dataset}

    Each training example is curated to be \textit{consistent with the Bradley-Terry model} of pairwise preferences.

    Specifically, for each ordered preference triplet \( (\tau^{(1)}, \tau^{(2)}, \alpha) \in \Xi \), the preference strength \( \alpha \in [0, 1] \) is computed based on the softmax-derived probability that \( \tau^{(2)} \) has a higher cost than \( \tau^{(1)} \) under the ground-truth costmap \( C_T \). Since $\tau^{(1)}$ is more preferred (lower cost) than $\tau^{(2)}$, the range of this probability is [0.5, 1] which we remap to [0,1] for \strength{}.
    
    We define:
    \[
    \alpha = 2 \left( \frac{\exp[\mu_{s^{(2)}}(C_T)]}{\exp[\mu_{s^{(1)}}(C_T)] + \exp[\mu_{s^{(2)}}(C_T)]} \right) - 1
    \].
    
\begin{figure}[h!]  % 'h' places the figure approximately here
    \centering
    \includegraphics[width=.5\textwidth]{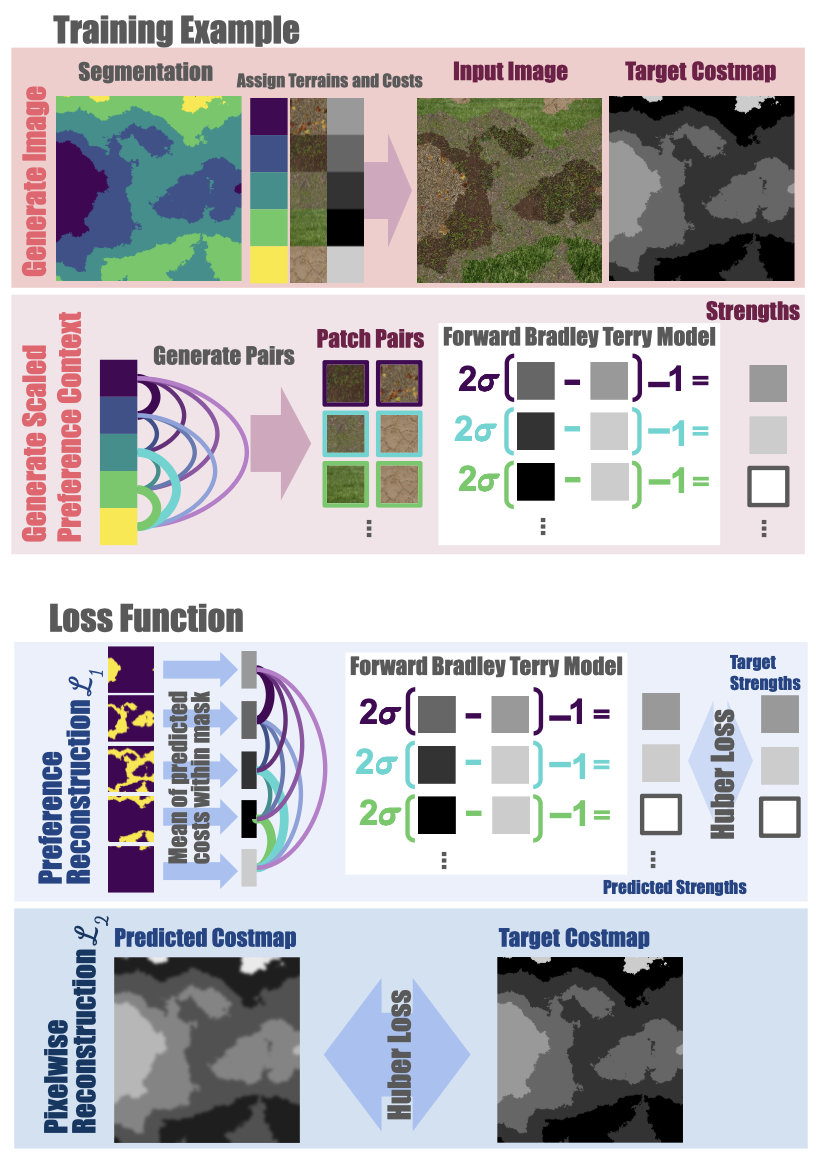}
    \caption{The construction of the Target Costmap and Scaled Preference Context in the Training Example parallels the loss function. The loss function has two parts: \textit{preference reconstruction} loss $\mathcal{L}_1$ and pixelwise reconstruction $\mathcal{L}_2$ against the target costmap. In $\mathcal{L}_1$, the preference context is constructed from the \textit{predicted} costmap using the ground truth segmentation masks, and the \textit{predicted} strengths are matched to the input (target) strengths. The $\mathcal{L}_2$ term keeps costs from exploding while the $\mathcal{L}_1$ term is critical for the model's preference-alignment ability. Note: $\mathcal{L}_1$ is shown in an equivalent form here for clarity, but we implement it as in eq. \ref{eq:loss1}.
    }
    \label{fig:training_example_loss_function}
    \vspace{-3mm}
\end{figure}

    To train and evaluate our model, we procedurally generate a synthetic dataset consisting of terrain images, ground-truth costmaps, and preference contexts. The dataset is built using a bank of pre-defined terrains and scalar cost values, combined using spatial masks.  We provide an ablation study on real versus synthetic data in Section \ref{sec:ablations}. The full generation process is outlined below (with steps 5,6 shown in Fig \ref{fig:training_example_loss_function}).

    \begin{enumerate}
        \item Define a bank of terrain types \( \terrains{} = \{\tau_1, \tau_2, \ldots, \tau_n\} \) and global scalar costs \(G = \{\gamma_1, \ldots, \gamma_m\} \).
        
        \item Generate \( k \) segmentation masks \( S = \{s_1, \ldots, s_k\} \) using procedural generation (e.g., the diamond-square algorithm \cite{diamond_square}).
        
        \item Sample \( k \) terrains \( \{\tau_1, \ldots, \tau_k\} \subset \terrains{} \) and \( k \) cost values \(\{\gamma_1, \ldots, \gamma_k\} \subset G \).
        
        \item Assign each terrain and cost to a mask, forming triplets \( \{(s_1, \tau_1, \gamma_1), \ldots, (s_k, \tau_k, \gamma_k)\} \).
        
        \item Use these assignments to construct the input image \( I \) and the ground-truth costmap \( C_T \).
        
        \item From the assigned triplets, construct a scaled preference context:
        \[
        \Xi = \left\{ \left(\tau_i^{(1)}, \tau_i^{(2)}, 2\sigma (\gamma_i^{(2)} - \gamma_i^{(1)})-1 \right) \right\}_{i = 1}^K
        \]
        
        \item Return \( I \), \( \Xi \), \( S \), and \( C_T \).
    \end{enumerate}

    The dataset contains 12 images of terrains ranging from \texttt{sand, mud, forest floor, river pebbles, grass, snow,} and \texttt{concrete} among others.
    % \gw{recommend using texttt for the class names to clue the reader into the fact that they're classes}
    Crucially, the use of a pretrained variational autoencoder (VAE) trained on internet-scale data allows the model to already encode strong priors about the structure and statistics of realistic natural images. As a result, our model can learn effectively from a relatively small number of synthetic BEVs, since the VAE provides a perceptual space aligned with the distribution of real-world visual inputs. This is helpful since size of the dataset is combinatorial with the number of BEV images.

% \subsection{Global Consistency for Aerial Maps}

% For large-scale deployment and aerial planning, the aerial maps are often too large for inference in a single pass, and we must tile the maps instead. Unlike \textsc{pacer} where there was an underlying ``global cost" given to each terrain during training, global consistency is not an objective during the training of \ourapproach{}. If only a single terrain is present in the local view, any cost would technically be consistent with the preference context. To rectify this underdeterminism, we input the image tiles at different scales \cite{Image_pyramids} and  take the mean across the output costmaps. We use tile scales of 1, $\frac{1}{3}$, and $\frac{1}{5}$ of the image size.

\section{Experiments}
To evaluate \ourapproach{}, we seek to answer the following questions:
 \textbf{(Q1)} Does \ourapproach{} generate accurate costmaps up to scale consistent with operator input?
 \textbf{(Q2)} Do paths planned over costmaps generated by \ourapproach{} outperform competing methods in terms of Hausdorff distance to ground-truth paths and the introduced metric of path \textit{regret}, which measures additional traversal cost compared to optimal paths? \textbf{(Q3)} How do our design choices of loss function and synthetic training data affect performance?

\lm{\textbf{Unfortunately, no other method operates on the same space of inputs}, so we provide classifier methods with more information--absolute costs for all classes--which users may not have in actual deployments.}

We use different datasets and baselines for \textbf{(Q1)} and \textbf{(Q2)} due to their distinct goals. \textbf{(Q1)} evaluates whether \ourapproach{} generates accurate costmaps up to scale, which requires ground-truth terrain labels. Since large-scale aerial datasets lack such pixel-level annotations—especially in off-road settings—we use the RELLIS-3D dataset \cite{jiang2021rellis}, which includes labeled terrain classes, and compare against segmentation-based baselines and the zero-shot method \textsc{pacer}. In contrast, \textbf{(Q2)} focuses on the planning utility of the generated costmaps in real-world, large-scale environments. Large-scale planning demands global aerial maps, which are typically unlabeled and unsuitable for supervised training. Therefore, we evaluate planning performance using these maps and compare against pretrained, open-set foundation models capable of generalizing without task-specific fine-tuning.

Since the costs from \ourapproach{} are correct up to scale, we normalize the local costmap $C$ by matching the lowest and highest predicted cost to the lowest and highest ground truth cost for all experiments.

\subsection{Accuracy of Generated Costmaps}
 Exploring \textbf{(Q1)}, we evaluate \ourapproach{} on the RELLIS-3D dataset \cite{jiang2021rellis} with three preferences $P$, and report the MAE to show how closely the varying preferences are adhered to. 

    \textbf{Methods and Baselines:}
    We compare against against \textsc{pacer} and dense segmentation models. First, since \textsc{pacer} does not support direct encoding of relative preference strengths, we approximate its behavior by providing only ordered terrain label pairs. Additionally, \textsc{pacer} requires a training phase on real-world data to learn a reasonable prior, due to its limited context length for expressing preferences. In contrast, \ourapproach{} is trained entirely on synthetic data (see Section~\ref{sec:synth_dataset}) and does not require task-specific fine-tuning at deployment time.
    The second baseline is a supervised segmentation model built on a pretrained \textsc{DeepLabV3\_ResNet50} backbone. We train two variants of this model: one fine-tuned on the RELLIS-3D dataset and the other on the same synthetic dataset used to train \ourapproach{}. The model trained on RELLIS-3D represents an upper-bound baseline, as it is trained and evaluated on the same domain. The synthetic-trained version provides a direct comparison point for \ourapproach{} under equivalent data conditions. While both methods use the same training data, \ourapproach{} significantly outperforms the segmentation model trained on synthetic data. This performance gap highlights the advantage of preference-driven costmap inference over traditional semantic segmentation, particularly in scenarios where semantic classes may not align with terrains traversed.
    
\begin{figure}
    \centering
    \vspace{-2mm}
    \includegraphics[width=0.48\textwidth]{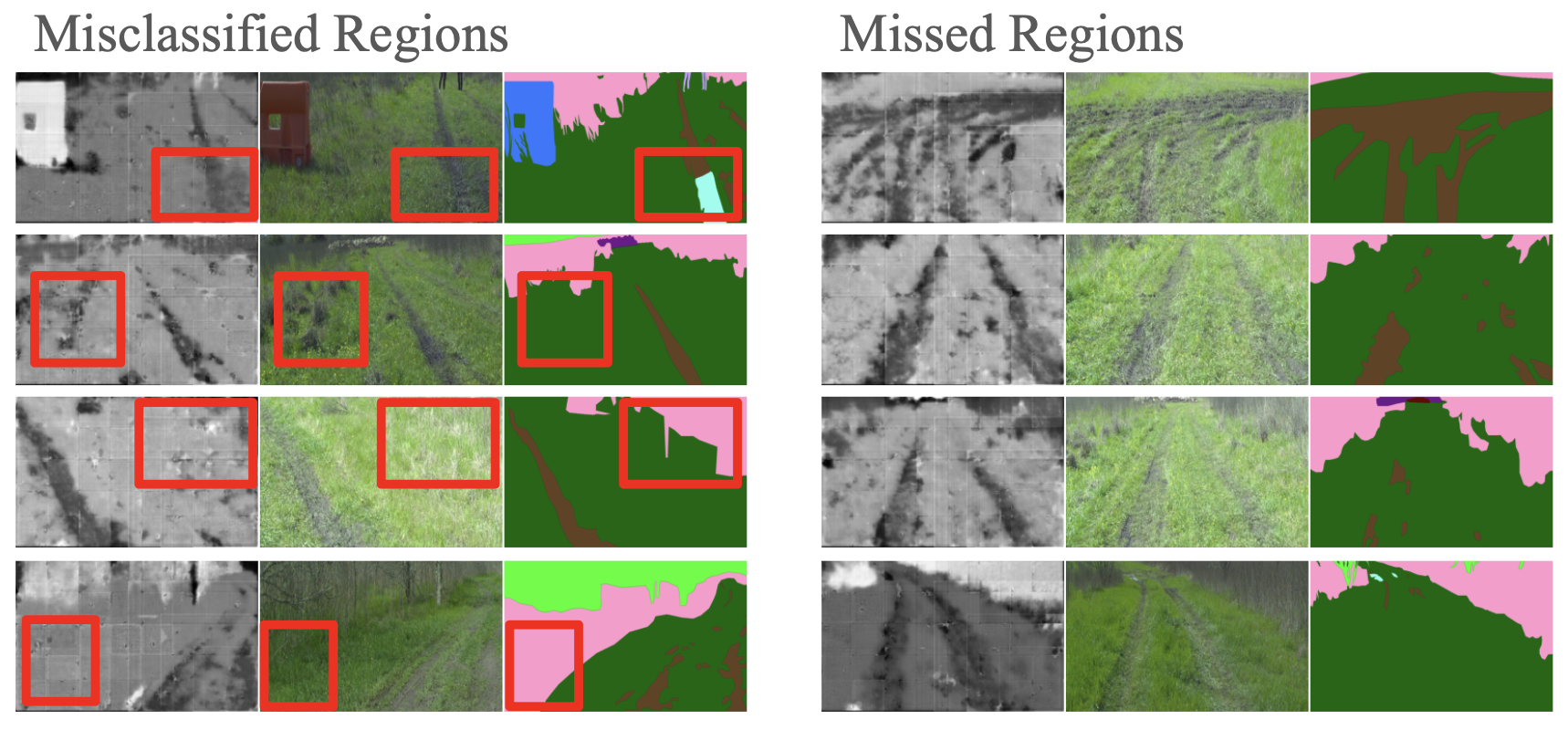}
    \caption{Examples of misclassified segmentation masks (left) and missed regions in segmentation masks (right). Each triplet consists of a \ourapproach{} costmap, RELLIS-3D image, and RELLIS-3D segmentation mask. We find that the RELLIS-3D segmentation masks are not accurate and fine enough for a quantitative evaluation across the whole dataset.}
    \label{fig:rellis_misclassifications}
    \vspace{-3mm}
\end{figure}

\textbf{Experimental Setup:} As we find there are misclassifications in many images (shown in Fig. \ref{fig:rellis_misclassifications} and futher discussed in Section \ref{sec:result}), we provide an evaluation over a subset of 860 images (cropped to remove the sky, and resized). \textbf{Metrics:} We report the mean absolute error (pixel-wise difference) per costmap for each method evaluated on three terrain class orderings, as well as the class-wise MAE for the three most frequent classes in Tab. \ref{tab:mse-results}.

\begin{table*}[tb]
\vspace{2mm}
\centering
\caption{Mean Absolute Error (lower is better) for each method under different preferences across selected semantic classes. Best is bolded. Second best is italicized. Asterisk * denotes the method is given \textit{more} information.}
\label{tab:mse-results}
\resizebox{\textwidth}{!}{
\begin{tabular}{lcccccc}
\toprule
\textbf{Preference} & \textbf{Method} & \textbf{Total MAE} & \textbf{Grass MAE} & \textbf{Bush MAE} & \textbf{Mud MAE} & \textbf{Barrier MAE} \\
\midrule

\multirow{5}{*}{P1} 
  & Segmentation (Ground Truth) *       & {0.0372} & {0.0197} & {0.0429} & {0.1724} & {0.1270} \\
  \hdashline
  & \ourapproach{}             & \textbf{0.1055} & \textbf{0.0711} & \textbf{0.1724} & \textbf{0.2215} & \textbf{0.0021} \\
  & \textsc{pacer}             & \textit{0.1374} & \textit{0.1041} & \textit{0.2059} & \textit{0.2734} & \textit{0.3105} \\
  % & \textsc{Sterling}            & \text{NaN} & \text{NaN} & \text{NaN} & \text{NaN} & \text{NaN} \\
  & Segmentation *    & 0.3003 & 0.2970 & 0.2732 & 0.3178 & 0.7677 \\

\midrule
\multirow{5}{*}{P2} 
  & Segmentation (Ground Truth) *       & {0.0359} & {0.0205} & {0.0424} & {0.1699} & {0.0323} \\
  \hdashline
  & \ourapproach{}             & \textbf{0.1035} & \textbf{0.0860} & \textbf{0.1193} & \textit{0.2218} & \textbf{0.0000} \\
  & \textsc{pacer}             & \textit{0.1611} & \textit{0.1571} & \textit{0.1545} & \textbf{0.2142} & 0.3451 \\
  % & \textsc{Sterling}            & \text{NaN} & \text{NaN} & \text{NaN} & \text{NaN} & \text{NaN} \\
  & Segmentation *    & 0.2914 & 0.3004 & 0.2700 & 0.3168 & \textit{0.3210} \\

\midrule
\multirow{5}{*}{P3} 
  & Segmentation (Ground Truth) *       & {0.0256} & {0.0140} & 0.0289 & 0.1296 & 0.0996 \\
  \hdashline
  & \ourapproach{}             & \textbf{0.1412} & \textbf{0.0501} & \textbf{0.0360} & \textbf{0.0466} & \textbf{0.0007} \\
  & \textsc{pacer}             & \textit{0.1661} & \textit{0.1853} & \textit{0.0791} & \textit{0.0653} & \textit{0.0768} \\
  % & \textsc{Sterling}            & \text{NaN} & \text{NaN} & \text{NaN} & \text{NaN} & \text{NaN} \\
  & Segmentation *    & 0.2951 & 0.2716 & 0.3607 & 0.3162 & 0.3050 \\
\bottomrule
\end{tabular}
}
\footnotesize Class distribution: Grass = 68\%, Bush = 18\%, Mud = 5\%, Barrier = 2\%
\vspace{-3mm}
\end{table*}

\textbf{Results \lm{and Discussion}:}
The Segmentation (Ground Truth) baseline performs best overall (Table \ref{tab:mse-results}), as it was trained directly on a subset of the RELLIS-3D dataset. The remaining three methods exhibit higher MAE due to discrepancies between the dataset’s provided ground-truth segmentation masks and the actual terrain textures in the images. Among these, \ourapproach{} consistently achieves the lowest total MAE and frequently the lowest class-wise MAE across all three evaluated preference orderings. 
\textsc{pacer} shows slightly higher MAE—while it is also preference-aware, its limited preference context length cannot fully capture all relevant comparisons for the many terrain classes present in RELLIS-3D. As a result, it leans more on learned priors over terrain and preference distributions, occasionally causing severe errors on underrepresented classes not shown above. Segmentation performs worst, constrained by the ontology of the synthetic dataset it was trained on. Unlike \ourapproach{}, it fails to generalize terrain understanding or preference modeling to the RELLIS-3D domain. \lm{These findings largely match our expectations: representation-learning methods (i.e. \textsc{pacer}) cannot modulate costs as flexibly and classification-based (i.e. segmentation baseline) methods cannot handle visually out-of-distribution terrains}. \textbf{Answering (Q1)}, we find that \ourapproach{} generates accurate costmaps that preserve relative cost magnitudes based on the given preference ordering and strengths.

\subsection{Accuracy of Planned Paths}
\label{sec:planning_experiments}

% \lm{d) The major problem of the paper: there is no discussion section in order to logically analyze the results, and assess if the results worked as predicted or not, what went well, what didnt go well, why, etc. }

In this experiment, we explore \textbf{(Q2)}: whether \ourapproach{} produces planned paths that better adhere to operator preferences than methods that ignore preference weights or require known classes.

\begin{figure}[t]
    \centering
    \vspace{-2mm} \includegraphics[width=0.5\textwidth]{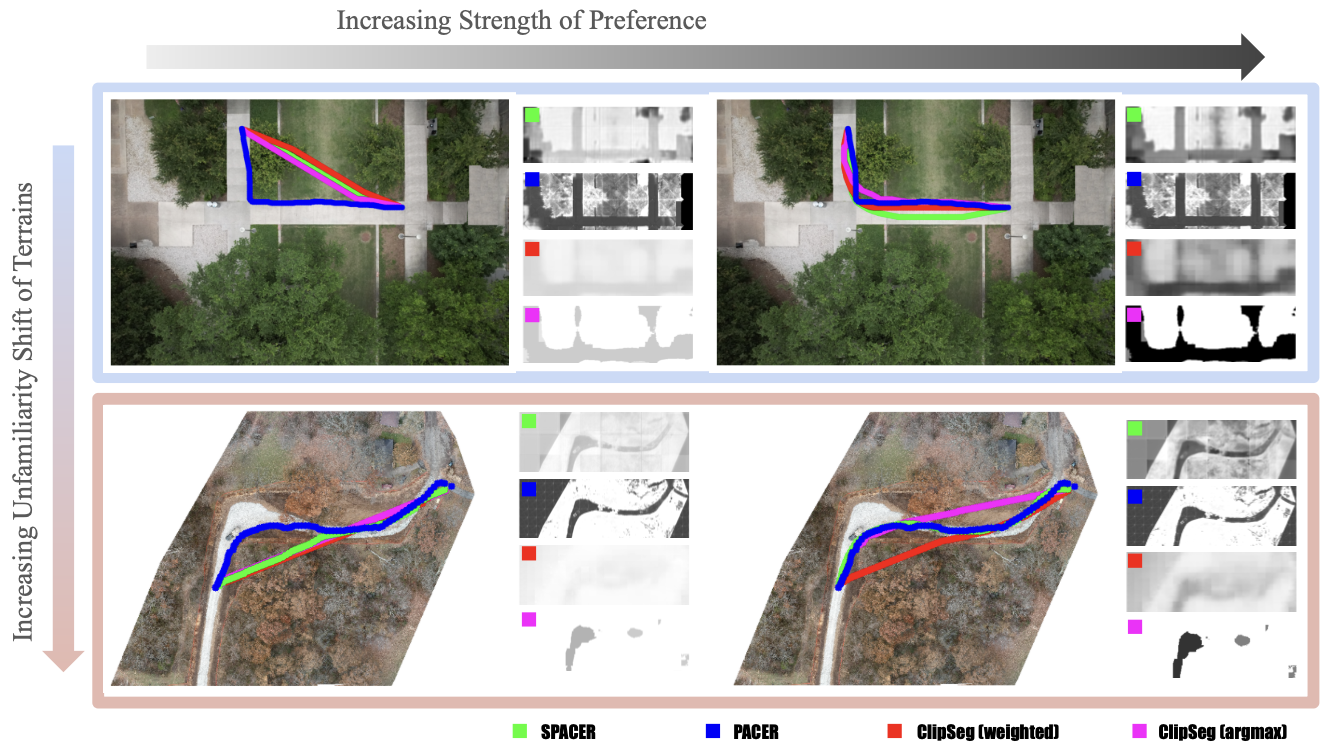}
    \caption{Qualitative comparison of costmaps generated by all methods. As preference strength varies across rows, \textsc{pacer} cannot adapt. As terrain \textit{visual} appearances become more misaligned with their classes across columns (i.e. vegetation is not green), classification baselines cannot adapt. \ourapproach{} produces costmaps and planned paths which are most aligned with the userspreferences}
    \label{fig:baselines_comparison}
    \vspace{-3mm}
\end{figure}

\textbf{Methods and Baselines:}
We evaluate \ourapproach{} against the following baselines: \textsc{pacer}, ClipSeg (weighted), and ClipSeg (argmax). ClipSeg (weighted) takes the linear combination of cost assignments to labels, weighted by the activation of the class label while ClipSeg (argmax) assigns cost based only on the most probable label. Both of these methods of open-set cost assignment are common, and each have limitations. ClipSeg (weighted) tends to produce costmaps that are smoother but less spatially precise in complex environments, while ClipSeg (argmax) yields sharper boundaries but can lead to high-variance costmaps. Qualitative examples are shown in \ref{fig:baselines_comparison}.% (Fig. \ref{fig:qualitative_paths})

% \lm{replace this figure with one of all the baselines}
% \begin{figure*}[t]
%     \centering
%     \vspace{2mm} \includegraphics[width=0.9\textwidth]{figures/aerial_vary_strength.png}
%     \caption{Varying Strength between Road and Lawn in an Aerial Map for Global Planning Experiments. As strength increases, the contrast in costs for these terrains increases.
%     }
%     \label{fig:strength_comparison_aerial}
%     \vspace{-3mm}
% \end{figure*}

\textbf{Experimental Setup:} We implemented a simulator which performs A* planning on aerial drone maps with a 40-connected angular lattice, where each node considers 40 discretized heading directions to enable fine-grained trajectory generation. This planner is fixed across all baselines to ensure comparability.
In each environment, we deploy the robot with a fixed terrain ordering in their preference, but vary the distances between the costs of terrains (i.e. vary the strengths). \textbf{Metrics:}  We evaluate each method using the Hausdorff distance between planned paths on predicted vs.\ ground truth costmaps, and two forms of \textit{regret}. \lm{The Hausdorff distance captures spatial deviation between paths, while regret quantifies suboptimality in cost}. Let $\bar{\Gamma}$ and $\Gamma$ denote the paths planned on ground truth and predicted costmaps, respectively, with cost functions $\bar{C}(\cdot)$, $C(\cdot)$. Then $\regret{}^* = \bar{C}(\Gamma) - \bar{C}(\bar{\Gamma})$ and $\hat{\regret{}} = C(\bar{\Gamma}) - C(\Gamma)$ represent \textit{regret} on the ground truth and predicted costmaps respectively. $\regret{}^*$ measures how optimal the predicted path is relative to the ground truth path, while $\hat{\regret{}}$ reflects how well the model evaluates the ground truth path under its own predicted costmap. We report the mean Hausdorff distance, $\regret^*$, and $\hat{\regret}$ across deployments for each environment. 

% The \ourapproach{}-generated global costmaps with varying strengths of Environment 7 are shown in Fig. \ref{fig:strength_comparison_aerial}.

\begin{table}[t]
\centering
\caption{Hausdorff Distance (lower is better) between planned paths on ground truth and predicted costmaps. Best value is bolded and second-best is italicized. Asterisk * denotes the method is given \textit{more} information. E1–E3 contain visually canonical terrains; E4–E7 contain non-visually canonical terrain.}
\label{tab:env-performance-hausdorff}
\resizebox{\linewidth}{!}{
\begin{tabular}{lccccccc}
\toprule
\textbf{Method} & \textbf{E1} & \textbf{E2} & \textbf{E3} & \textbf{E4} & \textbf{E5} & \textbf{E6} & \textbf{E7} \\
\midrule
\ourapproach{}      &  \emph{14.78} & \textbf{5.48} & \textbf{7.89} & \emph{14.96} & 44.99 & \emph{68.02} & \textbf{40.37} \\
\textsc{pacer}        & 23.21 & 22.84 & 19.37 & 20.26 & \emph{37.24} & \textbf{30.22} & 161.19 \\
ClipSeg (weighted) *    & \textbf{10.73} & 13.53 & \emph{10.02} & 51.21 & \textbf{30.05} & 100.77 & 193.55 \\
ClipSeg (argmax) *      & 19.78 & \emph{13.29} & 46.60 & \textbf{14.83} & 51.00 & 386.34 & \emph{64.57} \\
\bottomrule
\end{tabular}
}
\end{table}

\begin{table}
\centering
\caption{Performance evaluated on $\regret{}^*$ (lower is better). E1–E3 contain visually canonical terrains; E4–E7 contain non-visually canonical terrain.}
\label{tab:env-performance-regret-star}
\resizebox{\linewidth}{!}{
\begin{tabular}{lccccccc}
\toprule
\textbf{Method} & \textbf{E1} & \textbf{E2} & \textbf{E3} & \textbf{E4} & \textbf{E5} & \textbf{E6} & \textbf{E7} \\
\midrule
\ourapproach{}        & \textbf{1.40} & \textbf{1.05} & \textbf{0.90} & \textbf{3.77} & 25.50 & \textit{16.19} & \textbf{7.46} \\
\textsc{pacer}        & \textit{9.55} & \textit{5.81} & 2.82 & \textit{4.46} & \textbf{6.80} & \textbf{7.71} & 71.67 \\
ClipSeg (weighted) *    & 12.78 & 6.23 & 3.83 & 33.52 & \textit{13.29} & 302.44 & \textit{60.15} \\
ClipSeg (argmax) *      & 9.56 & 8.13 & \textit{1.78} & 40.66 & 71.58 & 241.27 & 63.43 \\
\bottomrule

\end{tabular}
}
\vspace{-3mm}
\end{table}

\vspace{0.2cm} 

\begin{table}
\centering
\caption{Performance evaluated on $\hat{\regret{}}$ (lower is better). E1–E3 contain visually canonical terrains; E4–E7 contain non-visually canonical terrain.}
\label{tab:env-performance-regret-hat}
\resizebox{\linewidth}{!}{
\begin{tabular}{lccccccc}
\toprule
\textbf{Method} & \textbf{E1} & \textbf{E2} & \textbf{E3} & \textbf{E4} & \textbf{E5} & \textbf{E6} & \textbf{E7} \\
\midrule
\ourapproach{}        & 10.27 & \textbf{2.03} & \textit{2.70} & \textit{15.39} & 56.80 & 81.75 & \textbf{17.16} \\
\textsc{pacer}        & 13.84 & 14.33 & 9.50 & 22.32 & 110.02 & \textit{61.18} & 87.15 \\
ClipSeg (weighted) *    & \textbf{3.21} & \textit{2.54} & \textbf{1.15} & 15.58 & \textit{11.08} & \textbf{53.84} & \textit{26.41} \\
ClipSeg (argmax) *      & \textit{3.78} & 3.60 & 3.41 & \textbf{14.00} & \textbf{9.72} & 212.92 & 30.78 \\
\bottomrule

\end{tabular}
}
\vspace{-3mm}
\end{table}

\textbf{Results \lm{and Discussion}:}
We find that \ourapproach{} achieved the lowest Hausdorff distance in two environments and the second-lowest in another three environments, for the overall best performance, as shown in Table \ref{tab:env-performance-hausdorff}. From Table \ref{tab:env-performance-regret-star}, we find that \ourapproach{} achieved the lowest $\regret{}^*$ in five out of the seven environments. \lm{\textsc{pacer} performs better on $\regret{}^*$ in E5 and E6, which we attribute to environment-specific variation or sensitivity to initialization rather than a systematic shortcoming of \ourapproach{}.} \ourapproach{} also achieved the lowest $\hat{\regret{}}$ in three environments.

\textsc{pacer} generally exhibited high $\hat{\regret{}}$ because it cannot adjust relative terrain costs, as shown in \ref{tab:env-performance-regret-hat}. As a result, it often considers the ground-truth path to be suboptimal when it traverses mildly unpreferred terrains, even though doing so is more efficient in terms of path length.

Open-set segmentation models showed high $\regret{}^*$, particularly in out-of-distribution (OOD) environments E4–E7. These models tend to struggle in novel settings due to mismatches between their learned visual concepts and the actual appearance of terrain in unfamiliar domains. 
% For instance, in one of our urban test environments featuring bright yellow-painted sidewalks, the models misclassified the terrain due to their reliance on canonical color and texture priors.
This often led to low-confidence predictions or misclassification, causing the planner to choose suboptimal routes that traverse undesirable terrain and incur high regret over longer trajectories.

Interestingly, ClipSeg (weighted) had consistently low $\hat{\regret{}}$ across all environments, likely because its output maps were low-confidence and blurry—effectively assigning similar costs throughout. As a result, the model evaluated the ground-truth path as only marginally worse than its own, despite producing poor predicted paths.

 \textbf{Answering  (Q2)}, we find that the preference-expression and generalizability of \ourapproach{} does lead to better planned paths than the baselines on a diverse set of aerial maps from different environments.

\subsection{Ablations}
\label{sec:ablations}

To explore \textbf{(Q3)}, we train our model on fully real (from robot deployments around a college campus) or fully synthetic data, using variations of our loss functions. We report Mean Absolute Error against the ground truth costmap. Each model is trained on five deployments and evaluated on three held-out deployments in different environments.

\begin{figure}
    \centering
    \vspace{-1mm}
    \includegraphics[width=0.4\textwidth]{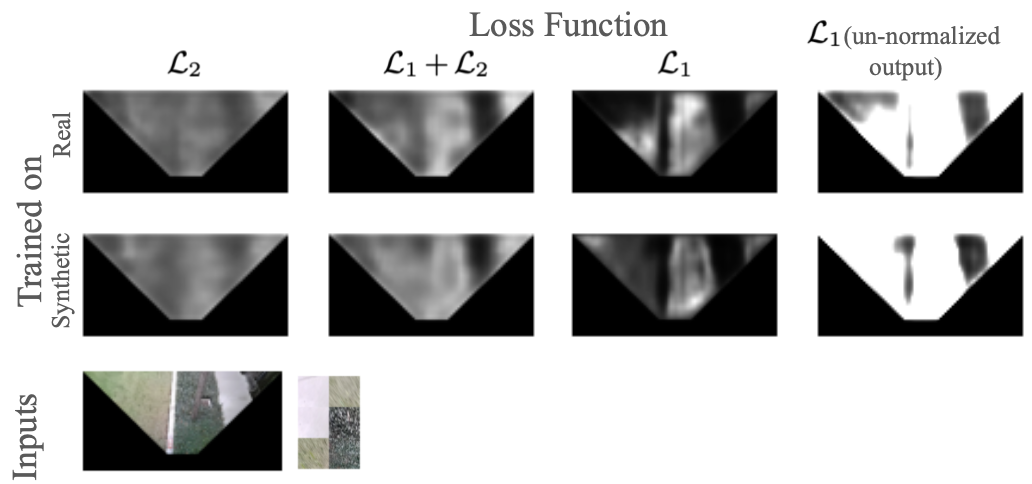}
    \caption{Qualitative performances of various models, comparing models trained with Real vs. Synthetic data and different loss functions.}
    \label{fig:ablation_qualitative_examples}
\end{figure}

Fig. \ref{fig:ablation_qualitative_examples} shows results from the ablation study. Rows represent realistic or synthetic training data, and columns represent the loss functions. In Fig. \ref{fig:ablation_qualitative_examples} and Tab. \ref{tab:loss_comparison}, we include both the un-normalized and normalized (divided by the maximum) output from the $\mathcal{L}_1$-only model in our results, since the raw output values have large magnitudes. Training only on pixel-wise reconstruction losses such as Huberloss (as is done in some prior work \cite{pacer}) results in poor performance, as it gives too weak a signal. The models trained only on $\mathcal{L}_1$ appear to follow the given preferences, though training is unstable and cost magnitudes tend to explode as the model is unregularized. Combining the losses, we weight each such that the magnitude of $\mathcal{L}_1$ is 5x greater than that of $\mathcal{L}_2$. The combined losses allow the model to learn human preferences while keeping output costs small and stable.

 Further, we find that training on purely synthetic data is not detrimental compared to real training data, with synthetic data being much less expensive to obtain.
 % \lm{The ground truth segmentation masks we train on are noisy and imperfect}. 
 Thus, we choose to train on synthetically-trained models with $\mathcal{L}_1 + \mathcal{L}_2$ loss.
 
% \lm{train with reconstruction loss only or $\mathcal{L}_1$ only and give results. the model with reconstruction loss only should fail. this shows that the new loss function introduced in this paper gives a much better signal than reconstruction (used in pacer 1)}.

\setlength{\tabcolsep}{2pt} 
\begin{table}[tb]
\centering
\begin{tabular}{lcccc}
\toprule
\textbf{Training Data} & \textbf{$\mathcal{L}_2$} & \textbf{$\mathcal{L}_1 + \mathcal{L}_2$}& \textbf{$\mathcal{L}_1$} & \makecell{\textbf{$\mathcal{L}_1$} \textbf{(un-normalized)}} \\
\midrule
Real Training Data      & 0.1516 & \textbf{0.1157} & 0.1350 & 1.0965 \\
Synthetic Training Data & 0.1459 & \textbf{0.1205} & 0.1391 & 1.2124 \\
\bottomrule
\end{tabular}
\caption{Mean Absolute Error (lower is better) averaged across three robot deployments in new environments, comparing performances of various models trained with Real vs. Synthetic data and different loss functions. $\mathcal{L}_1 + \mathcal{L}_2$ performs best, with little between \textit{real} and \textit{synthetic} data.}
\label{tab:loss_comparison}
\vspace{-3mm}
\end{table}

\subsection{Runtimes}
\lm{Inference time for a single forward pass was benchmarked on an NVIDIA RTX A6000 GPU. \textsc{pacer} achieved the fastest runtime at 0.012 s per image, followed by \textsc{DeepLabV3\_ResNet50} (0.029 s), \ourapproach{} (0.057 s), and CLIPSeg (0.058 s).}

%===============================================================================

\section{Further Qualitative Results}
\label{sec:result}

Using \ourapproach{}, we found that many images in the RELLIS-3D dataset either have coarse segmentations or misclassified regions. We provide examples below of contention between the terrains and labeled masks. In addition to this uncertainty regarding where the segmentation boundaries should lie, we find that terrains within the same class label can vary widely in appearance (Fig. \ref{fig:rellis_misclassifications}). These are limitations with segmentation approaches in general, so we provide a limited quantitative evaluation of this dataset as well as a qualitative evaluation, the latter of which we claim is more important for evaluating performance.

% In Fig. \ref{fig:strength_comparison}, we show that the predicted local costmap is conditioned on the strengths given in the preference context, when the pairs are kept constant. When the strength is low (close to zero) the regions of the costmap corresponding to these terrains are similar shades of gray. When the strength is high (close to 1), the regions of the costmap corresponding to these terrains have a stronger contrast in cost.

\begin{figure}[tb]  % 'h' places the figure approximately here
    \centering
    \vspace{1mm}
    \includegraphics[width=.5\textwidth]{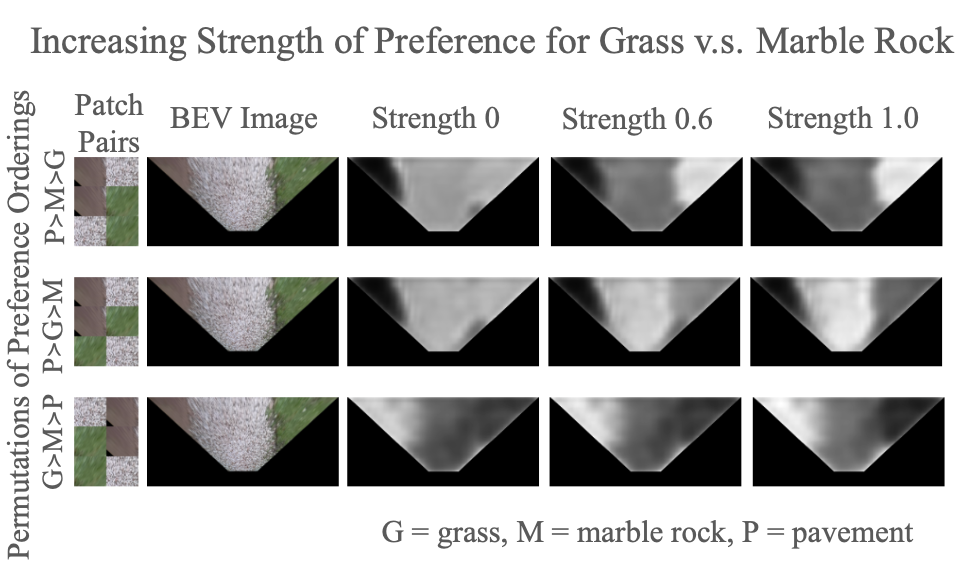}  % Replace with your filename
    \caption{Effects of increasing strength of preference of grass and marble rock (3rd pair in the context) versus preference orderings over terrains. %\jb{This is too small to read.}
    }
    \label{fig:strength_comparison}
    \vspace{-3mm}
\end{figure}

Qualitative results from varying preference orderings and strengths for a BEV image (collected during a robot deployment on a university campus) is shown in Fig. \ref{fig:strength_comparison}.  Each row shows a different total ordering over three terrains  (\texttt{pebble pavement}, \texttt{marble rock}, 
\texttt{grass}) as represented by the patch pairs. Each column of costmaps shows the model output at a different strength of the \texttt{grass} v.s. \texttt{marble rock} pair. The other two pairs have a strength of 1. At low strengths, grass and marble rock are given similar costs. At high strengths, there is strong contrast between the costs for grass and marble rock. In all cases, the cost ordering is consistent with the total ordering dictated by the patch pairs.
    % \subsubsection{Varying the Preference Context Length}

    % \subsubsection{Conflicting Preferences}
%===============================================================================

% \section{Inferring Strengths}
% \label{sec:inferring_strengths}

% A numerical strength may be difficult for an inexperienced human user to provide. We therefore present a prototype interface for calculating strengths from other formats of preference expression. 
% The user may simply click on terrain patches to form an ordered preference context. Given a planner, the system iteratively guesses and refines the strength parameters through a binary search, according to the user's feedback. At each step, we sample several parameter values and have the user select the best planned path so far (Fig. \ref{fig:overview_fig}). The parameters are iteratively adjusted for several rounds. Following this period of parameter-tuning, the full preference context (including strengths) may be used for full deployment. Here, we have shown that \ourapproach{} can be used with human input but without the need for the human to give numerical values.
% % \jb{This is an interesting method! Does it work if planning over more than two terrain types?} \lm{uh maybe someone else can get it to work with more than two terrain types}

\section{Conclusion}
\label{sec:conclusion}
This paper introduced \ourapproach{}, a novel framework for zero-shot terrain-aware navigation that enables fine-grained user control over terrain cost magnitudes in novel, out-of-distribution environments. By addressing the limitations of fixed semantic ontologies and ingrained visual priors in existing segmentation and representation-learning approaches, \ourapproach{} introduces a more expressive and flexible formulation of user preferences. Our method leverages a new mathematical representation of preference intensity, a targeted synthetic data generation process, and a tailored loss function to train models that are both generalizable and responsive to user intent. In future work, we aim to extend \ourapproach{} beyond RGB-only input to include modalities such as depth or multimodal signals.
% \commentp{Need at least a sentence or two of limitations/future work.}

%===============================================================================
\begingroup
\scriptsize
\bibliographystyle{IEEEtran}
\bibliography{example}  % .bib
\endgroup

\end{document}